\documentclass{article}
\usepackage{ijcai21}

\usepackage[T1]{fontenc} 
\usepackage{booktabs} 
\usepackage{amsfonts} 
\usepackage{nicefrac} 
\usepackage{microtype} 

\usepackage{times}
\usepackage{soul}
\usepackage{url}
\usepackage[hidelinks]{hyperref}
\usepackage[utf8]{inputenc}
\usepackage[small]{caption}
\usepackage{graphicx}
\usepackage{amsmath}
\usepackage{amsthm}
\usepackage{booktabs}
\usepackage{algorithm}
\usepackage{algorithmic}

\usepackage{natbib}
\usepackage{url}

\usepackage{subcaption}

\usepackage{todonotes}

\usepackage{multirow}
\usepackage{bbm}

\usepackage{wrapfig}

\usepackage{mathtools}

\DeclareMathOperator*{\argmax}{arg\,max}

\usepackage{xr}

\title{Learning Interpretable Concept-Based Models with Human Feedback}

\author{
Isaac Lage$^1$\footnote{Contact Author}\and
Finale Doshi-Velez$^1$\\
\affiliations
$^1$John A. Paulson School of Engineering and Applied Sciences\\Harvard University\\Cambridge, MA, 02138\\
\emails
\texttt{isaaclage@g.harvard.edu},
\texttt{finale@seas.harvard.edu}
}

\begin{document}

\maketitle

\begin{abstract}
Machine learning models that first learn a representation of a domain in terms of human-understandable concepts, then use it to make predictions, have been proposed to facilitate interpretation and interaction with models trained on high-dimensional data. However these methods have important limitations: the way they define concepts are not inherently interpretable, and they assume that concept labels either exist for individual instances or can easily be acquired from users. These limitations are particularly acute for high-dimensional tabular features. We propose an approach for learning a set of transparent concept definitions in high-dimensional tabular data that relies on users labeling \textit{concept features} instead of individual instances. Our method produces concepts that both align with users' intuitive sense of what a concept means, and facilitate prediction of the downstream label by a transparent machine learning model. This ensures that the full model is transparent and intuitive, and as predictive as possible given this constraint. We demonstrate with simulated user feedback on real prediction problems, including one in a clinical domain, that this kind of direct feedback is much more efficient at learning solutions that align with ground truth concept definitions than alternative transparent approaches that rely on labeling instances or other existing interaction mechanisms, while maintaining similar predictive performance.
\end{abstract}

\section{Introduction}

Concept-based machine learning methods express predictions in terms of high-level concepts derived from raw features instead of in terms of the raw features themselves. In general, the goal is for machine-learned concepts to align with human users' internal concepts (rather than simply being representations that increase the machine's predictive performance). Because the concepts are intuitive to users, they can facilitate interpretation and interaction with models in a way not possible at the level of the high-dimensional raw features. For example, \citet{Kim2018CAV} uses concepts to understand biases in neural networks trained on image data, and \citet{koh2020ConceptBottleneck} argues that models with a concept bottleneck layer can be intervened on to make correct predictions even when concept predictions are incorrect.

In this work, we focus on methods to identify concepts from the kinds of high-dimensional, tabular, count data that frequently occur in healthcare records. In these settings, the raw data consist of quite granular codes (e.g. patient has codes for Lorazepam and generalized anxiety), and the clinician's mental model operates at a higher level of patient condition (e.g. patient has anxiety). Methods to turn these raw features into concepts that clinicians can easily reason about, and then explain predictions in terms of these concepts, should be easier to understand and manipulate than those expressed in terms of raw diagnostic codes. 

However existing concept-based models have limitations that make them difficult to apply in settings like the clinical one described above. The first major limitation is that these methods define concepts in terms of black-box classifiers, so they may fail to accurately capture the user's mental model of the concept in any of the number of ways that black box models may fail (spurious correlations, lack of fairness, etc.), and without tools to interpret \textit{the concepts}, a user's attempt to interpret the predictive model based on them may also fail. The second major limitation is that these methods rely on access to ground-truth concept labels for at least some fraction of the dataset. However, in many cases this annotation can require significant user effort. This is exacerbated by the fact that the process of chart review to obtain a patient condition label from health data is much more time-consuming than labeling an image, which is the setting that many previous works have considered.

\begin{figure}[h!]
 \centering
 \includegraphics[width=0.9\linewidth]{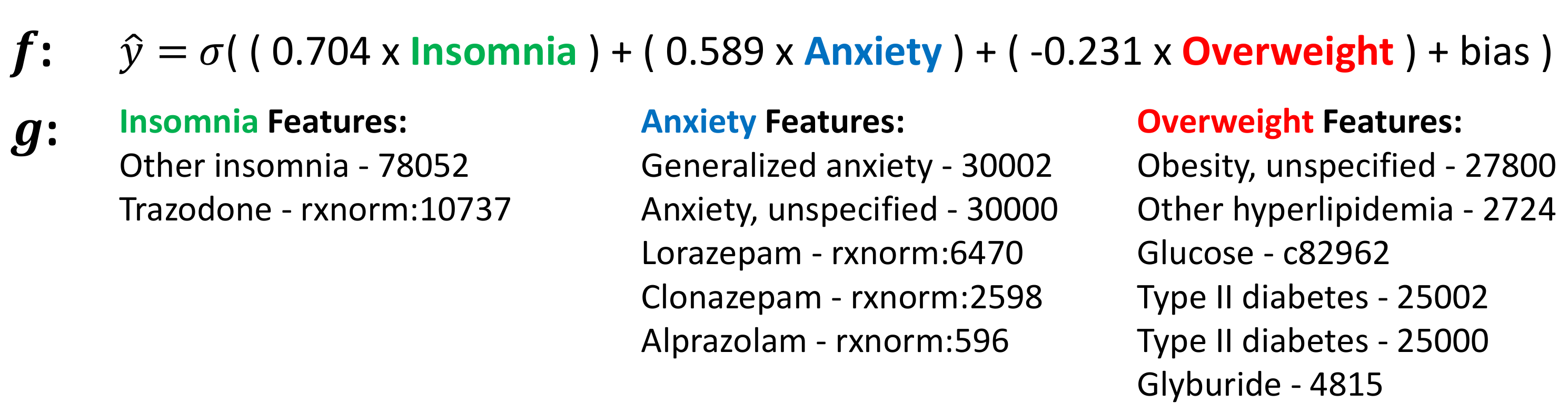}
 \caption{An example model of the form used by our method in a clinical domain. The $g$ component is a transparent representation layer that generates intuitive concepts, and the $f$ component is a transparent model learned on top of the representation.}
 \label{fig:learned_concepts}
\end{figure}

We propose an approach to learning concept-based models that addresses both of these limitations. We learn a concept-based model where the concepts are fully transparent, thus enabling users to validate whether mental and machine models are aligned. Our learning process also incorporates user feedback directly when learning the concept definitions: rather than labeling data, users mark whether specific \textit{feature dimensions} are relevant to a concept. This process further enhances human-machine concept alignment by allowing the user more control over the learned concept, and makes efficient use of the user's time, particularly since \citet{raghavan2006features} found that labeling a feature takes a fraction of the time of labeling an instance. We present a procedure for maximizing the model's predictive performance while adhering to these constraints on interpretability, and the alignment of the concept with the user's internal concepts. (See Figure~\ref{fig:learned_concepts} for examples of concept definitions of the form proposed in our method.) We demonstrate with simulated user feedback on real prediction problems, including one in a clinical domain, that this kind of direct feedback is much more efficient at learning solutions that align with ground truth concept definitions than alternative transparent approaches that rely on labeling instances or other existing interaction mechanisms, while maintaining similar predictive performance.

\section{Related Work}

\paragraph{Transparent Machine Learning.} Transparency has been proposed as one instantiation of interpretability corresponding to whether a user can step through a model's computation in a reasonable amount of time \citep{lipton2016interpretability}. It can help users avoid pitfalls of black box models like unfairness, relying on spurious correlations, and a variety of other errors \citep{rudin2019stop}. Many machine learning models have been proposed to satisfy this criteria including sparse logistic regression (\citet{tibshirani1996lasso}, \citet{poursabzisangdeh2018manipulatingmeasuring}), decision sets (\citet{lakkaraju2016}, \citet{lage2019humanevaluation}) and rule lists (\citet{ustun2016slim}, \citet{letham2015rulelists}). However these approaches are presented in terms of raw features, assuming they are meaningful. Our work aims to develop a transparent representation that maps uninterpretable features in high-dimensional domains to an intuitive and transparent concept representation on top of which existing transparent machine learning methods can be used. 

\paragraph{Uninterpretable Concept-Based Models.} Approaches have been proposed to give intuitive meaning to latent spaces of complex (uninterpretable) models. Most closely related to ours, \citet{koh2020ConceptBottleneck} introduces an approach for constructing a bottleneck layer in a neural network that corresponds to a set of concepts that are labeled in the training data. They demonstrate that these concepts can then be manipulated to change and understand the model's output instead of directly manipulating input features. Other related lines of work use semi-supervised approaches where a user provides some labels for latent dimensions in order to constrain the latent space to align with the user's representation of the problem (\citet{Narayanaswamy2017DisentangledDGM}, \citet{Hristov2018LatentSpaceDemonstration}), or interpret the latent space of a neural network in terms of intuitive concepts post-hoc by allowing users to specify concepts in terms of examples that train a classifier on a neural network's latent space \citep{kim2017interpretability}. In contrast to this, our method learns a model that is fully transparent, not post-hoc, and does not require a dataset of instances labeled with concepts. 

\paragraph{Interactive Concept Learning.} Approaches to interactively learn a concept-based representation have been proposed, several of which produce interpretable concept-based representations. \citet{amershi2009concept} introduces ``end-user interactive concept learning," through which users can generate labels for relevant concepts in large datasets to train concept classifiers, which can be done with transparent classifiers. 
Interactive topic models have been proposed to obtain low-dimensional representations aligned with user intuition; these are linear, positive and can be interpreted by the top features for each topic. 
Mechanisms for interaction include constraints that words should or should not appear in the same topic, \citep{hu2014interactivetopic}, and defining a set of ``anchor words" to characterize a topic; the latter is considered easier to guide. \citet{lund2018classifier} builds on it to learn predictive topics for downstream prediction tasks. Finally, \citet{parikh2011attributes} learns mid-level features for image classification by jointly finding predictive hyper-planes, and learning a model to predict the nameability of those hyper-planes, but this depends heavily on users being able to inspect instances to see how a latent feature varies among them. This works well with images, but it is not clear how this would work with data that is not glaneceable. While these methods allow for interactively learning interpretable concepts, the feedback users provide does not allow them to directly steer the algorithm, making it challenging for users to get the concepts to align with their intuitive representation of the problem.

\paragraph{Interactive Feature Engineering.} Additional methods have been proposed to allow users to interactively engineer and refine feature spaces. \citet{cheng2015hybridml} and \citet{takahama2018adaflock} use crowd-workers to label images with features generated by other crowd-workers, then iteratively refine the feature space for examples that are frequently misclassified. Some active learning methods have been proposed that allow features to give feedback on the relevance of features rather than labeling instances e.g. \citep{raghavan2006features, druck2008features}. In user studies, \citet{raghavan2006features} and \citet{druck2008features} demonstrated that people are able to label features with their relevance to a prediction task, and \citet{raghavan2006features} demonstrated that labeling features actually takes much less time than labeling instances. In contrast to our method, these all aim to increase predictive performance of the downstream model with user feedback, rather than to tune the model to be intuitive to the user.

\section{Interactively Learning Intuitive Concepts}

Our goal is to predict label $y \in \{0, 1\}$ given input feature vector of count data: $x \in \mathcal{Z^+}^D$. We assume access to a dataset $\{x_n, y_n\}^{N}$ and aim to learn a 2-stage prediction function first mapping from input features $x$ to concepts $c \in \{0, 1\}^C$--we call this the concept definition, $g$, and then mapping concepts $c$ to predictions $\hat{y}$--we call this the prediction function, $f$. 

Our goal is to learn an $f$ and $g$ that are as predictive as possible, while learning an $f$ that is interpretable, and a $g$ that is both interpretable and intuitive, in that it closely corresponds to a set of concepts the user is already familiar with. This gives us the following objective function that we solve:
\begin{equation}
 \label{eqn:objective}
	\begin{aligned}
	 \argmax_{f, g} \quad \texttt{accuracy}(f(g(x)), y) \\ 
	\text{subject to\quad} f \text{ is interpretable} \\
	g \text{ is interpretable} \\
	g \text{ is intuitive}
	\end{aligned}
\end{equation}
Below, we instantiate $f$ and $g$ which have functional forms that are transparent by design. Then we introduce an optimization procedure to fit the parameters of $f$ and $g$ so that they have the correct functional form, $g$ is intuitive--i.e. aligned to the user's internal concept definition, and $(f(g(x)))$ is predictive. Section~\ref{sec:proposals} describes a core piece of the optimization framework in more detail.

\subsection{Interpretable Concept Definition $c = g(x)$}

The key claim underlying our method is that, if the concept definitions $g$ produce concepts that closely align with the user's internal concepts, then the entire predictive model consisting of the functions $f$ and $g$ can be interpreted by only inspecting $f$. This reduces the effort required to interpret the model. To accomplish this, we require that concept definition $g$ be interpretable so that the user can validate that it produces concepts that closely match their own. Without this requirement, problems that interpretability aims to solve, such as models learning spurious correlations, can creep into $g$ and influence the interpretation of the entire model.

To ensure that the learned concept definitions $g$ are interpretable, we draw inspiration from the medical literature where conditions are often manually defined from high dimensional record data. One common form is defining a condition (one possible instantiation of a concept) based on a threshold on a sum of counts. This form of concept definition is known to be interpretable to humans since it is the de-facto clinical approach (e.g. \citet{castro2015bipolarphenotype}, \citet{townsend2012depression}, \citet{ritchie2010genotype}).

We formalize a simplified version of this popular form without a count threshold by defining the concept definition $g$ using a binary matrix of parameters, $A \in \{0, 1\}^{D\texttt{x}C}$ where setting $A_{i,j} = 1$ indicates that feature $i$ is associated with concept $j$. A prediction $c_{j}$ can then be made for concept $j$ as follows:
\begin{equation}
 \label{eqn:concepts}
 c_j = \mathbbm{1}( ( A_j x ) \geq 1 )
\end{equation}
Features $i$ where $A_{i,j} = 1$ form a list of features that are associated with concept $c_j$. The main goal of our approach will be to learn this set of features.

\subsection{Interpretable Prediction Function $\hat{y} = f(c)$}

Since our goal is to interpret the model in terms of the concepts instead of the raw input features, the prediction function $f$ that depends on the concepts must be interpretable. In this work, we shall use logistic regression, but in general, any differentiable and interpretable model could be used. Let $W \in \mathcal{R}^C$ be the vector of weights and $b \in \mathcal{R}$ the scalar bias. The prediction can be written:
\begin{equation}
 \label{eqn:prediction-function}
 \hat{y} = \sigma(W^T c + b)
\end{equation} The model can then be interpreted in terms of the concept weights, $W$.

\subsection{Optimization procedure}

Given the forms we have defined for $f$ and $g$, we must define a procedure for solving our objective in Equation~\ref{eqn:objective}. The main challenge to our optimization is the last constraint: that $g$ must be intuitive. Since intuitive concepts are a property of the user's understanding of a particular domain rather than a property of the data itself, we must assess this property with the help of the user.

We define an optimization procedure that relies on being able to query the user about whether a feature $x_i$ should be associated with concept $c_j$. For example, an association between the feature `generalized anxiety' and the concept `anxiety' might make sense, whereas an association with `anxiety' and `other insomnia' does not---even though it might make the concept more predictive of the patient's psychiatric outcomes. We assume that if the user accepts associating feature $x_i$ with concept $c_j$ for every $(i, j)$ feature-concept association in $g$, then $g$ is intuitive and satisfies the constraint. I.e. the constraint is satisfied if the user accepts associating $x_i$ with $c_j$ $\forall \{(i,j): A_{i,j} = 1\}$.

To guarantee that we will learn a $g$ that satisfies this property, we first require the user to initialize $g$ with an intuitive association for each concept, grounding the concepts to something the user finds intuitive. Our algorithm then iteratively builds up $g$ by making a series of feature-concept proposals $(i^*, j^*)$ that the user must accept or reject. If the proposal is accepted, the feature-concept association is added to the concept definition, improving the quality of $g$ while guaranteeing that it satisfies the constraint, and if it is rejected, the concept definition does not change so $g$ continues to satisfy the constraint. 

We also make efficient use of human feedback by modeling which associations the user has previously accepted to refine future proposals made by the algorithm, and we refit $f$ after every change to $g$ to increase the overall predictiveness of the model. Each of these steps is described below, and the full algorithm is described in Algorithm~\ref{alg:algorithm}.

\paragraph{Initialization} The user first initializes the binary feature-concept association matrix, $A$, by specifying exactly one feature they wish to be associated with each concept: $\forall j \sum_{i=0}^D A_{(i, j)} = 1$.) In clinical domains like the Psych one we use in our experiments, clinicians already know many of the high-level concepts of interest that will affect the prediction, and coming up with a few examples of features related to these concepts (e.g. generalized anxiety) seems relatively straightforward, but specifying the long tail of relevant features (e.g. alprazolam, lorazepam etc.) can be challenging; this is where our approach is useful.

\paragraph{Proposals} The optimization then proceeds by, at each step, proposing a feature-concept pair, $(i^*, j^*)$ from the set of $\{(i, j)\}$ pairs that have not yet been explored by the algorithm. We call the unexplored set of features associated with concept $c_j$: $u_j$ (for unlabeled), and the previously explored set $l_j$ (for labeled). The user then gives feedback by either accepting the association, thus setting feature-concept matrix $A_{i, j} = 1$, or by rejecting the association, in which case no change is made to $A$. $l_j$ is initialized as $\{i: A_{i,j'} = 1\}$ for any concept $j'$, and $u_j$ is initialized as the complement of $l_j$. This is further described in Section~\ref{sec:proposals}.

\paragraph{Labels for intuitive associations} We additionally store a set of labels indicating where the user has previously accepted or rejected a proposal: $\texttt{intuit}$. These allow us to model which associations are or are not intuitive, and to refine the quality of future proposals. These are initialized so that $\texttt{intuit}_{i, j} = 1$ and $\texttt{intuit}_{i, j' \neq j} = 0$ if $A_{i, j} = 1$ in the concept definitions initialized by the user. We then update these intuitiveness labels setting $\texttt{intuit}_{i^*, j^*} = 1$ if proposal $(i^*, j^*)$ is accepted and $\texttt{intuit}_{i^*, j^*} = 0$ if it is rejected. In the initialization, we assume that features can only be associated with one concept, but we do not make this assumption for our proposals. This reflects the assumption that features used to seed the concepts will be more cleanly aligned with them than features suggested by our algorithm.

\paragraph{Additional details} After each accepted change to the feature-concept matrix $A$, we refit $f$ so that it is optimally predictive given the updated concept definitions. We fix the concept for each proposal, $j^*$, optimizing only over the feature dimension, $i^*$. We make a fixed number of proposals to the user for each concept before moving onto the next in order to minimize the mental load placed on the user by continuously switching between concepts. To short-circuit the loop of human feedback when no predictive proposals exist, we add a dummy feature that corresponds to making no change to the concept, and do not request feedback from the user when this is proposed. 

\begin{algorithm}[H]
	\begin{algorithmic}[1]
		\STATE{Input: $x$, $y$, $A$, $k$}
		\STATE{Initialize: $f$, $l$, $u$, $\texttt{intuit}$}
		\FOR{$j^*$ 1:num-concepts}
			\FOR{k 1:num-proposals}
				\STATE{Choose feature $i^*$ to construct proposal: $(i^*, j^*)$} \label{alg:line:proposal}
				\IF{$(i^*, j^*)$ is accepted}
					\STATE{$\texttt{intuit}_{i^*, j^*}=1$}
					\STATE{$A_{i^*, j^*} = 1$}
					\STATE{Retrain $f$}
				\ELSE
					\STATE{$\texttt{intuit}_{i^*, j^*}=0$}
				\ENDIF
				\STATE{$l_{j^*} = l_{j^*} \cup \{i^*\}; u_{j^*} = u_{j^*} \setminus \{i^*\}$}
			\ENDFOR
		\ENDFOR
	\end{algorithmic}
	\caption{Our algorithm for interactively optimizing intuitive and interpretable concepts with human feedback. See Section~\ref{sec:proposals} for details on how we choose feature $i^*$ when making proposals.}\label{alg:algorithm}
\end{algorithm}

\section{Proposing Predictive, Likely-Intuitive Changes}
\label{sec:proposals}

The key challenge in the optimization process outlined above is how to make proposals that will both increase the predictive performance, and will also likely satisfy the intuitiveness constraint--i.e. how to implement Line~\ref{alg:line:proposal} in Algorithm~\ref{alg:algorithm}. If the proposal is not highly predictive, it will not improve the objective much even if the user accepts it (and thus be a waste of the user's time), while if the proposal is not intuitive, that is, not aligned to the user's mental model, then the user will not accept it and no improvement will occur.

To produce proposals that are predictive and intuitive, we first compute both how predictive a particular feature-concept association is likely to be---we denote this $\texttt{score}^{\texttt{pred}}$---and how likely to user is to accept it---we denote this $\texttt{score}^{\texttt{intuit}}$. Next we combine both these scores---we denote the combiner function $h_k(\texttt{score}^{\texttt{pred}}, \texttt{score}^{\texttt{intuit}})$, to rank potential proposals. $k$ corresponds to a threshold parameter used by the combiner function. Below, we describe the procedures for computing $\texttt{score}^{\texttt{pred}}$ and $\texttt{score}^{\texttt{intuit}}$ before describing how they are combined in $h_k$.

\paragraph{Computing $\texttt{score}^{\texttt{pred}}$} 

The goal of $\texttt{score}^{\texttt{pred}}_{i, j}$ is to measure how much the predictive performance of the model will increase if feature $i$ is associated with concept $j$. We want to use this score to prioritize proposals that will improve the predictive performance as much as possible if the user decides to accept the proposal. 

We achieve this by updating the model to its form if the proposal $(i, j)$ were to be accepted by the user, and computing the predictive performance under this update (without retraining $f$ because of the computational cost). We denote this update $\tilde{A}^{i, j}_{i, j}$, and it is identical to $A$ except that $\tilde{A}^{i, j}_{i, j} = 1$. Formally, $\texttt{score}^{\texttt{pred}}_{i, j} = \texttt{accuracy}(y, f(g(x, \tilde{A}^{i, j}))$ for $i \in u_j$, and $0$ o.w. While brute force approaches like this can be computationally expensive, in this case, it allows us to compute the $\texttt{score}^{\texttt{pred}}_{i, j}$ quantity in only several seconds (each proposal for the Yelp domain took approximately 2 seconds on a single core of an Intel Core i7 processor). This appears to be within the bounds of reasonable behavior for an interactive system; \citet{lund2018classifier} report that participants were unable to complete a task when updates took more than 10 seconds.

\paragraph{Computing $\texttt{score}^{\texttt{intuit}}$} 

The goal of $\texttt{score}^{\texttt{intuit}}_{i, j}$ is to measure how likely the user is to accept associating a feature $i$ with a concept $j$. When this is high, we can propose a feature-concept pair with confidence, knowing that the user is likely to accept the change. When it is low, we should prioritize other proposals that the user is more likely to accept. 

We operationalize $\texttt{score}^{\texttt{intuit}}$ using a Gaussian random field (GRF) \citep{zhu2003} to model the user's probability of accepting a proposal. This model assumes that the user is likely to accept associating a feature $i$ with a concept $j$ if the user has previously accepted associating similar features $i'$ with concept $j$ and unlikely to do so if they have previously rejected similar features. This requires defining a notion of similarity between features: we use Jaccard similarity (denoted $J$) computed over the number of times each features is recorded for each instance (i.e. $x^T$). Synonymous features are likely to be used somewhat interchangeably throughout a patient's medical history, for example, making this notion of similarity reasonable.

The probability that the user will accept associating feature $i$ with concept $j$ can then be computed based on propagating the $\texttt{intuit}$ labels, denoting whether the user has previously accepted or rejected a proposal, through the similarity graph:
 \small
\begin{eqnarray}\label{eq_energy}
 \texttt{score}\overset{\texttt{intuit}}{_{i, j} = } \exp(\frac{1}{2} \sum_{i' \in l_j} J(x^T_i, x^T_{i'}) (\texttt{intuit}_{i, j} - \texttt{intuit}_{i', j})^2) \nonumber
\end{eqnarray}
\normalsize

\paragraph{Making Proposals: $h_k(\texttt{score}^{\texttt{pred}}, \texttt{score}^{\texttt{intuit}})$}

Our goal is produce a proposal of a feature-concept association, $(i^*, j^*)$, that will both be highly predictive, and that the user is likely to accept. Meeting both of these conditions makes it likely that it will both be added to the model, and have a positive effect on predictive performance. 

Recall that for a given proposal, the concept, $j^*$, is fixed. We compute the proposal by first taking the top $k$ features in $u_{j^*}$--the features not yet probed by the algorithm for concept $j^*$, ranked by $\texttt{score}^{\texttt{pred}}$. The intuition is that any of these are predictive enough to help with downstream performance if the user were to accept the proposal. We then take the highest ranking feature amongst these $k$ ranked by $\texttt{score}^{\texttt{intuit}}$. This corresponds to choosing the term that the user is most likely to accept from amongst the predictive features found in the first step. 

This thresholding and re-reranking allows us to incorporate both the association's predictiveness and the likelihood the user will accept it when making proposals. 

\section{Illustrative Toy Example}
\label{sec:toy-example}

We introduce a simple, toy example to illustrate a scenario where neither $\texttt{score}^{\texttt{pred}}$ (pred) nor $\texttt{score}^{\texttt{intuit}}$ (intuit) alone make high quality proposals, but their combination, $h_k(\texttt{score}^{\texttt{pred}}, \texttt{score}^{\texttt{intuit}})$ (our method, pred-intuit), can.

\begin{figure}[h!]
 \centering
 \includegraphics[width=.4\linewidth]{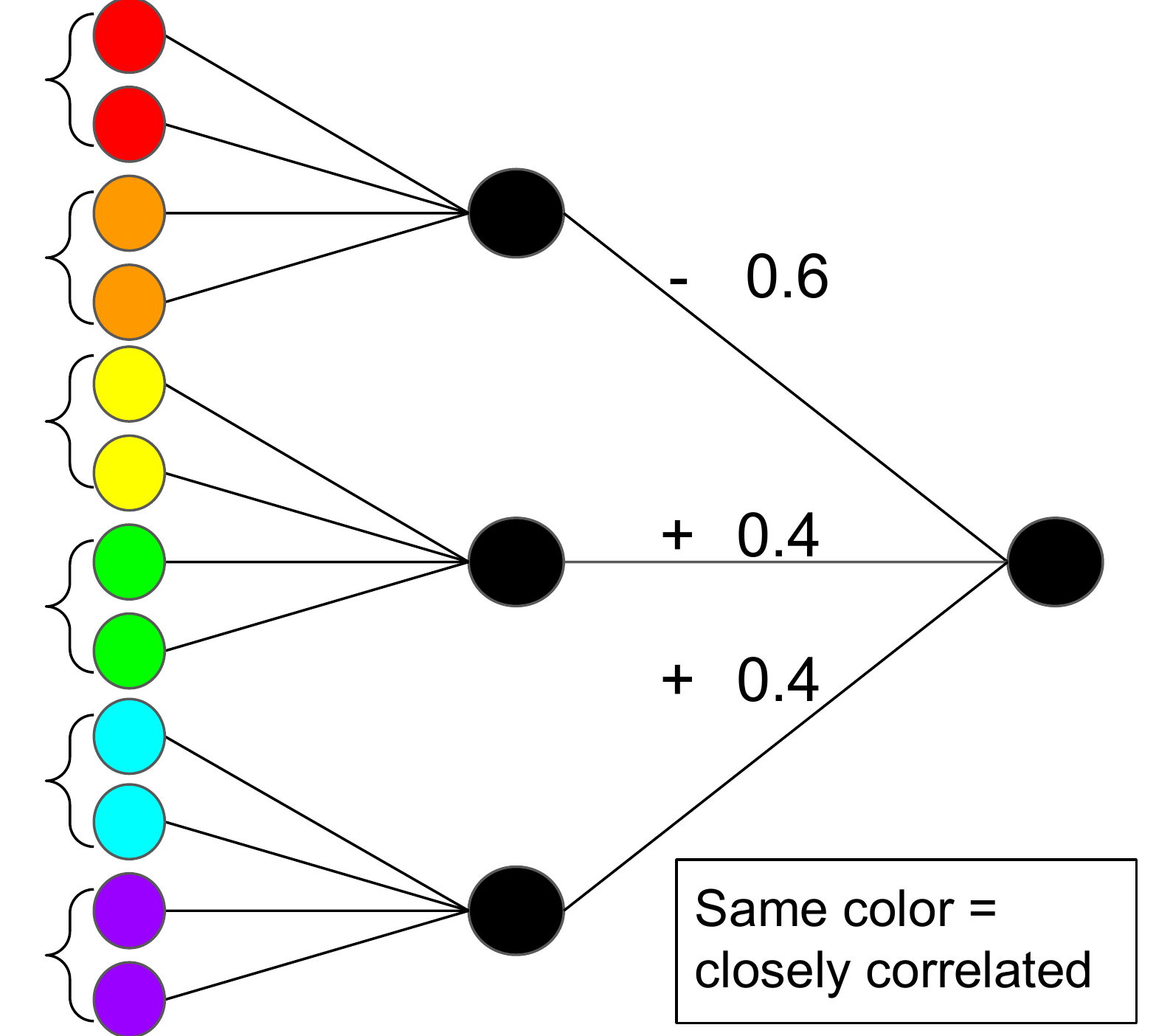}
 \includegraphics[width=.55\linewidth]{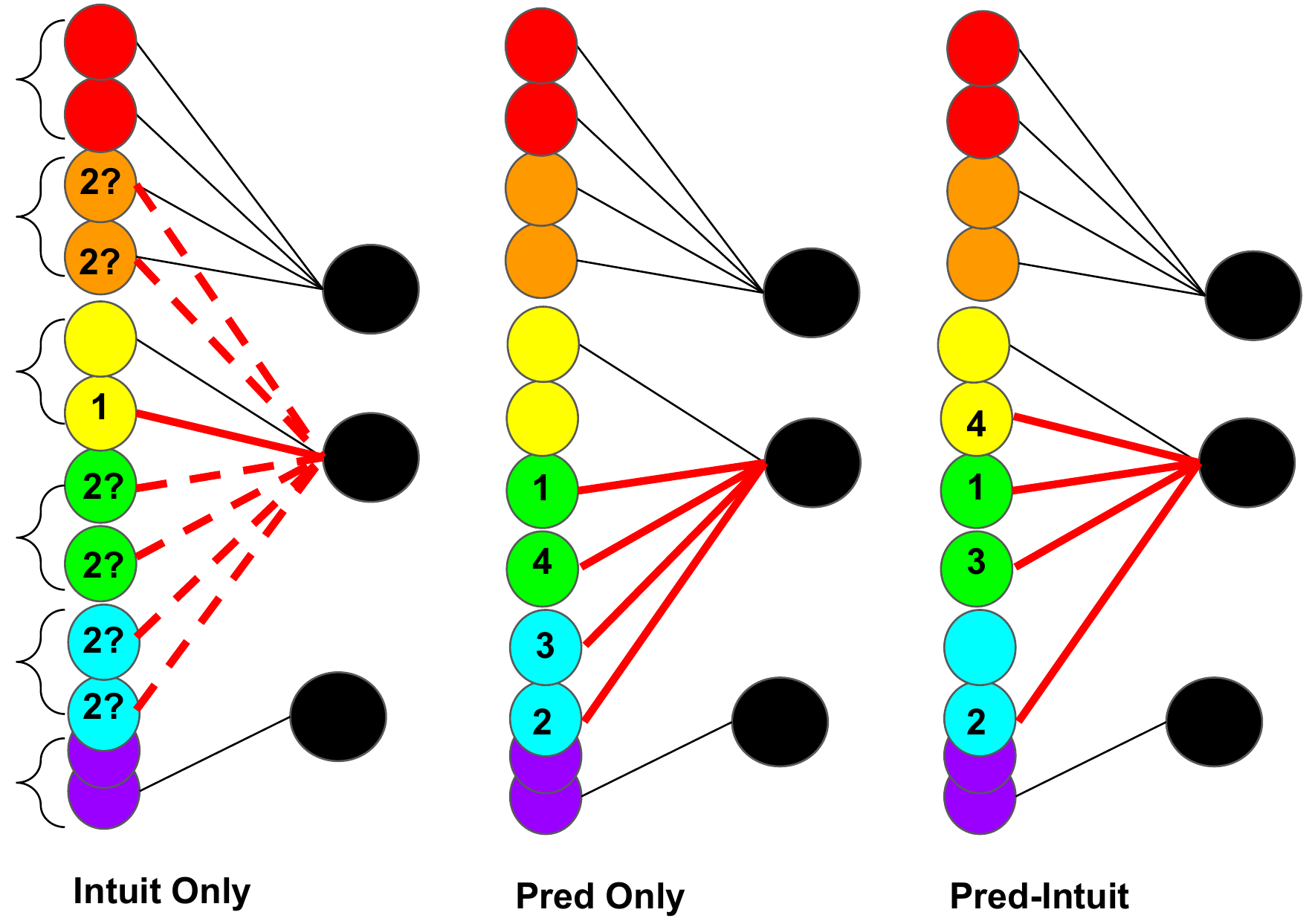}
 \caption{The toy example has 2 concepts that play similar roles in the prediction, and each concept depends on 2 distinct sets of correlated features. Black edges correspond to previously added associations, red edges correspond to the next proposals made by the algorithm, and dotted red edges indicate there are multiple best options for that proposal. The numbers in the corresponding feature nodes correspond to the ordering of the red proposals. The intuit variant can fill in the correlated sets of features, but is not effective at uncovering new sets of features, while the pred variant is unable to learn that features associated with a different concept that plays a similar predictive role are not relevant. Our approach (pred-intuit) avoids both of these pitfalls.}
 \label{fig:toy-example}
\end{figure}

\textbf{Properties} This example has 2 key properties that we expect to occur in real data. The first is that two distinct concepts have a similar effect on the prediction, which may be the case if, for example, both anxiety and insomnia have similar effects on psychiatric outcomes, even though they are 2 distinct clinical conditions. The second is that each concept consists of 2 groups of 2 highly correlated features, however these groups are not correlated. This may be the case when, a single condition may consist of 2 disjoint groups of patients, like patients with type I and type II diabetes, for example. Figure~\ref{fig:toy-example} shows the details of the toy example, where nodes that have the same color are closely correlated, edges for $g$ have weight 1 and edges for $f$ have their weight written on them. See Appendix Section~\ref{sec:dataset-appendix} for a detailed description of how the features were generated.

\textbf{Results} Figure~\ref{fig:toy-example} describes key failure cases for each variant. Black edges correspond to previously added associations, red edges correspond to the next 4 proposals made by the algorithm, and dotted red edges indicate there are multiple options for that proposal. The numbers in the corresponding feature nodes indicate to the ordering of the red proposals. The intuit variant is able to successfully add the 2nd yellow node at the start of the 2nd concept when the concept was seeded with the first yellow node because these are clearly correlated. However for its second proposal, it can add any of the green, cyan or orange nodes nodes, since none of these are correlated with either the seed term for concept 2, or for any other concept. The pred strategy first proposes the green node, since this correctly labels the most additional instances with the concept, then proposes both of the cyan nodes because concept 3 (that the cyan nodes are associated with) plays the same predictive role as concept 2. After exhausting the cyan nodes, it returns to the second green node, however does not learn to do so immediately after the first rejected proposal, as our method does. 

When run with 4 proposals for each concept, our method that combines both predictiveness and intuitiveness is able to recover the true set of features, and achieves almost perfect downstream and concept accuracy (concept accuracy: $0.994 \pm 0.001$, downstream accuracy: $0.990 \pm 0.001$), while the pred variant performs slightly worse on both metrics (concept accuracy: $0.978 \pm 0.001$, downstream accuracy: $0.966 \pm 0.001$), and the intuit variant performs even worse (concept accuracy: $0.892 \pm 0.015$, downstream accuracy: $0.858 \pm 0.018$). See Appendix Figure~\ref{fig:toy-results}.

\section{Experiments}

To allow for quantitative analysis and comparison to multiple baselines and variants of our approach, we ran experiments with known (hand-crafted) concepts to be discovered from real data: each experiment was seeded with features from the known concept, and we assumed that the simulated user would accept any proposed term that belonged to it.

\subsection{Setup}

\paragraph{Datasets and Concept Definitions}

We use two domains: one publicly available dataset of Yelp restaurant reviews\footnote{\url{https://www.yelp.com/dataset/}}, and one real, clinical dataset of patients diagnosed with depression from a Boston area hospital. In the Yelp data, we predict whether the average rating for a restaurant is good ($\geq$ 4 stars), or bad ($\leq$ 2 stars) based on counts of words in the aggregated reviews. In the Psych dataset, we predict whether a patient will be prescribed an atypical antipsychotic within 1 year of their first antidepressant prescription based on counts of the patient's past diagnoses, prescriptions and procedures. After preprocessing, the Yelp dataset has dimension $7,496\texttt{x}1,228$, and the Psych dataset has dimensions $9,802\texttt{x}989$; both are split 60/20/20 train/valid/test, and labels are class balanced by subsampling. Neither of these real datasets come with concept definitions, so we crafted these concepts by hand to be realistic and reasonably predictive. The concepts in the Psych domain were informed by discussions with a practicing psychiatrist. See Appendix~\ref{sec:dataset-appendix} for additional details.

\paragraph{Baselines}
We compare to interactive, concept-based baselines as well as non-concept-based predictors. The concept-based baselines are comparable to our $g$ (concept definitions), and have the same $f$ as our method trained on top of them to make predictions. Since both approaches are interactive, they require an interaction method through which the user can provide feedback on the quality of the concepts--this is analogous to the step in our method where we propose a feature-concept association and the user provides feedback on whether it is intuitive. The first baseline uses active learning to fit a sparse, l1-regularized logistic regression classifier for each concept--denoted AL. Here, the feedback from the user consists of labeling instances with concept labels (whether the concept applies to that instance). The second baseline uses a supervised variant of anchor-topic-modeling \citep{lund2018classifier}, where the user feedback indicates whether a feature should be an anchor for a particular topic--denoted TM. Anchors are features that have a non-zero probability of appearing in only one topic, and can be used to guide the learned topics to take on a specific meaning. See Supplement Section~\ref{sec:hyperparameters-appendix} for method details, hyperparameters, and results for an additional TM variant. 

For non-interactive baselines that do not rely on concepts, we compare to a random forest classifier (RF), a neural network with a single hidden layer the of size $C$ (NN), and l1-regularized logistic regression with a similar number of non-zero coefficients as $C$ (LR). See Supplement Section~\ref{sec:hyperparameters-appendix} for hyperparameters and details.

Finally, we explore variants of our approach: a pred variant that ranks proposals by $\texttt{score}^{\texttt{pred}}$, an intuit variant that ranks proposals by $\texttt{score}^{\texttt{intuit}}$, several variants of our method (`pred-intuit'), and several `intuit-pred' variants that threshold based on $\texttt{score}^{\texttt{intuit}}$ then rank based on $\texttt{score}^{\texttt{pred}}$. Different variants correspond to different choices of $k$: 1\%, 5\%, 10\%, and 25\% of D. We denote these as e.g. pred-intuit-5 when $k = 5\%$ of $D$. $k$ is not chosen by hyperparameter optimization since we do not have a validation set of concepts, instead we choose k=5\% (pred-intuit-5) a priori to show in our main results as this value of $k$ seems neither too restrictive, nor too broad.

\subsection{Comparison to Baselines}
The downstream accuracies and the concept accuracies on the test set are reported with standard errors in Table~\ref{tab:baselines} for 25 randomly sampled sets of seed features from the list in Appendix~\ref{sec:dataset-appendix} and 10 proposals per concept. 
 
\setlength{\tabcolsep}{3pt}
\begin{table}[h!]
\centering
\small
 \begin{tabular}{||l|r r|r r||} 
 	\hline
 	& \multicolumn{2}{|c|}{Yelp} & \multicolumn{2}{|c||}{Psych} \\
	Variant & Downstream & Concept & Downstream & Concept \\
	\hline\hline
	\textbf{Ours} & 0.780$\pm$0.003 & \textbf{0.707$\pm$0.002} & \textbf{0.606$\pm$0.002} & \textbf{0.778$\pm$0.010} \\
	AL & 0.753$\pm$0.003 & 0.612$\pm$0.003 & 0.581$\pm$0.003 & 0.598$\pm$0.013 \\
	TM & \textbf{0.811$\pm$0.009} & 0.596$\pm$0.004 & \textbf{0.611$\pm$0.003} & 0.648$\pm$0.004 \\ \hline
	LR & 0.696$\pm$0.000 & - & 0.588$\pm$0.000 & - \\ 
	NN & 0.909$\pm$0.002 & - & 0.635$\pm$0.002 & - \\ 
	RF & 0.935$\pm$0.001 & - & 0.670$\pm$0.001 & - \\ \hline
 \hline
 \end{tabular}
 \caption{Downstream accuracy and concept accuracy $\pm$ standard errors for our method and baselines in both domains on the heldout test set. Bold numbers indicate the best accuracy(ies) for the concept-based methods. Our method performs substantially better on concept accuracy than concept learning baselines, while staying competitive on downstream accuracy. All interpretable baselines (above the horizontal line) have worse prediction than blackbox regressors (below line).}
 \label{tab:baselines}
\end{table}
\setlength{\tabcolsep}{6pt}

\setlength{\tabcolsep}{3pt}
\begin{table*}[]
\centering
\tiny
\begin{tabular}{|lll|lll|lll|}
\hline
\multicolumn{3}{|c|}{TM}   & \multicolumn{3}{|c|}{AL} & \multicolumn{3}{|c|}{Ours} \\ \hline
C1 & C2 & C3 & C1 & C2 & C3 \\ \hline
\textbf{managers} 	& \textbf{neighborhood} & \textbf{creamy}	& drivethru				& \textbf{cozy}			& \textbf{creamy}		& \textbf{reservation}	&	\textbf{comfortable}	& \textbf{juicy}		\\
pull 				& \textbf{vibe} 		& \textbf{texture}	& nicely				& garbage				& macdonalds			&						&	\textbf{welcoming}		& \textbf{moist}		\\	
standing 			& playing				& fruit				& \textbf{reservation} 	& receipt				& shift					&						&	\textbf{vibe}			& \textbf{creamy}		\\
bought 				& watch					& vegas				& lovely				& macdonalds			& addition 				&						&	\textbf{casual}			&						\\
issues				& bartender				& \textbf{rich}		& entrees				& \textbf{comfortable} 	& perhaps 				&						&	\textbf{ambiance}		&						\\ \hline
\end{tabular}
 \caption{Top 5 coefficients for each concept in AL, TM and Our method in Yelp (from restart with highest concept accuracy). Bolded features are related to the concept they are associated with. While for AL and TM some of the top features are relevant, many are irrelevant to the concept. For ours, the top (and only) features associated with each concept are all relevant by design.}
 \label{tab:coefficients}
\end{table*}

\textbf{Our approach substantially outperforms all methods on concept accuracy.}
In Yelp, our final concept accuracy is $0.71\pm0.0$, and in Psych, it is $0.78\pm0.1$, $0.1$ and $0.13$ greater than the 2nd best concept-based model respectively. These substantial differences suggest that our approach aligns much better with the user's intuitive representation than baselines, given a fixed number of user interactions. This is crucial for deriving valid interpretations of the predictive model in terms of these concepts.

\textbf{Our approach is competitive with other concept-based approaches on downstream prediction accuracy.} 
Our approach outperforms AL in both domains, and performs similarly to TM in the real clinical domain. However TM outperforms our approach in the Yelp domain, which may simply be a property of the Yelp data, as topic models are particularly well-suited to text data. These results suggest that our approach consistently performs well, including on real clinical data, that it is more effective than the AL baseline at optimizing downstream performance, and that it does not lose much on downstream accuracy compared to the TM baseline.

\textbf{Our approach performs better on downstream accuracy than the LR baseline, which is comparable to training $f$ directly on the raw features.} In the Yelp domain, our approach outperforms LR by $0.08$, and in the Psych domain, by $0.02$. This suggests that training an interpretable model on top of the concepts constructed by our approach can boost predictive performance over training a sparse logistic regression on the raw features. Our concept based approach has other advantages as well, including that the predictors in the interpretable model are user specified, whereas the inputs to a sparse logistic regression do not have any constraint on intuitiveness, and that colinearity \citep{dormann2013collinearity} may cause weights on the raw features to be confused in e.g. sparse logistic regression, while the user chosen concepts are more likely to represent distinct ideas.

\textbf{Standard predictors without interpretability constraints outperform all concept-based approaches on downstream accuracy.} All the concept-based methods (including ours) have worse downstream accuracy than the non-interpretable methods (NN and RF); however, we emphasize that (a) neither of these baselines are interpretable and (b) there may be several ways to narrow that gap---the most substantial of which is moving beyond the particular concepts initialized by the user to find concepts that are both intuitive to the user, and highly predictive.

\begin{figure}[h!]
 \centering
 \includegraphics[width=.45\linewidth]{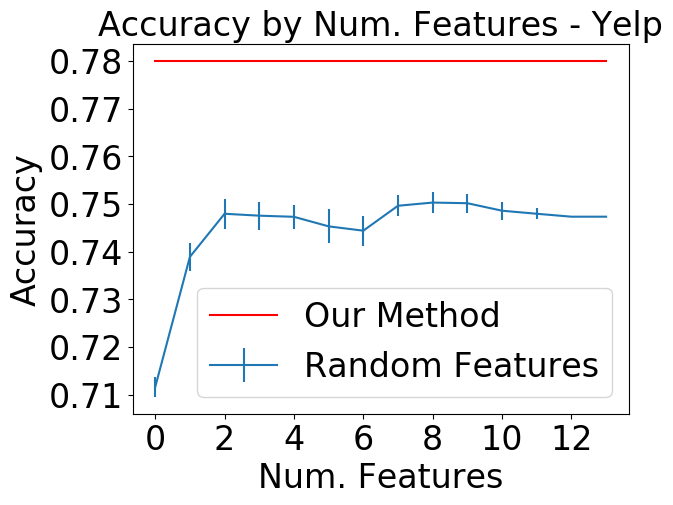}
 \includegraphics[width=.45\linewidth]{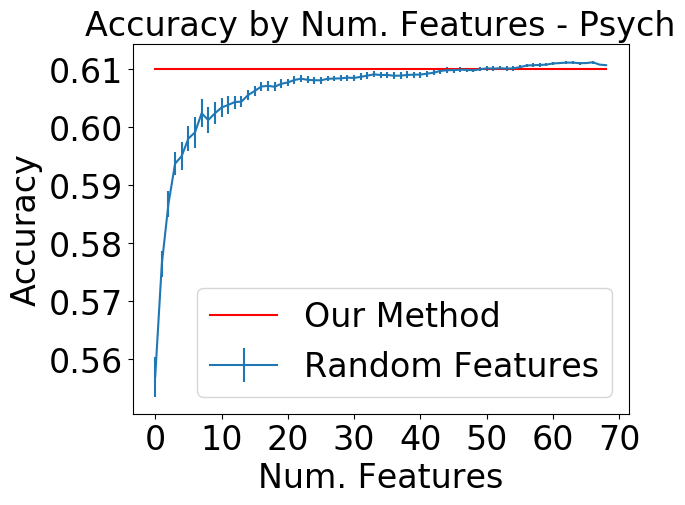}
 \caption{Heldout downstream accuracy by number of sampled relevant features. Yelp domain on left, Psych on right. In the Yelp domain, randomly adding features to the concepts from the list of ``ground truth'' related features never approaches the downstream performance of our method, while in the Psych domain, it takes up to around 50 random features per concept to surpass what our method achieves with 10 proposals per concept.}
 \label{fig:random-codes}
\end{figure}

\textbf{Comparisons against fully manual: Our approach outperforms the user manually selecting a small set of relevant features.} We compare the downstream accuracy of our approach on the test set against randomly sampling features from the concept definitions to simulate a user generating $g$ manually. The x-axis corresponds to the number of features sampled (without replacement) from each concept. Once all of the features for a given concept have been sampled, that concept is no longer updated. Figure~\ref{fig:random-codes} shows this for the 25 randomly selected sets of seed features used to generate Table~\ref{tab:baselines}. In Yelp, comparable downstream accuracy is never reached, and for Psych it takes approximately 50 features for some of the concepts to achieve comparable accuracy. The poor performance in the Yelp domain suggests that there are features that intuitively belong to the concept for the user, but that actively hurt predictive performance; our approach avoids adding these features. Overall, these results suggest that manually curating a predictive $g$ will require much more effort than seeding our approach with 1 relevant term. 

\textbf{Spurious correlations in baseline concepts: Baselines have many irrelevant features in the top features for each concept; our approach, by design, does not.} In Table~\ref{tab:coefficients}, we report the top 5 features associated with each concept in the Yelp domain for both the TM and AL baselines. We see that, some of them are relevant to the topic at hand, for example `vibe' for TM-C2, and `comfortable' for AL-C2. Other examples are bolded in the table. However many of these features are not directly associated with the concepts, for example, `vegas' in TM-C3, and `addition' in AL-C3. These may lead to errors that could be caught by inspecting the model since it is interpretable, but unlike our method, AL and TM are not designed to allow the user to easily correct these errors.

\subsection{Ablations and Variants} 

\begin{figure}[h!]
 \centering
 \includegraphics[width=0.5\textwidth]{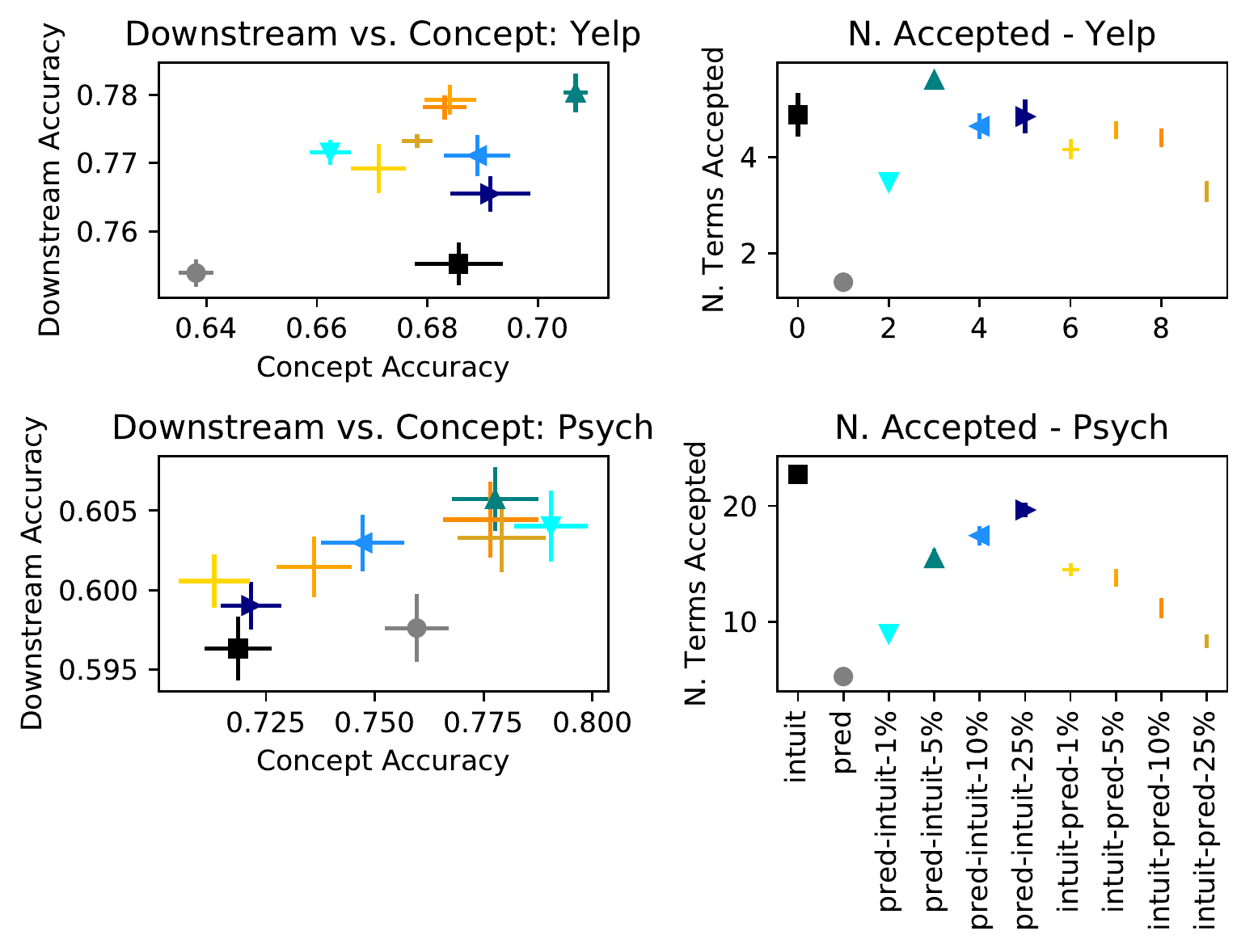}
 \caption{Results of our ablation study on the importance of both predictiveness and intuitiveness in the proposal. Variants that include both factors outperform pred and intuit in both concept accuracy and downstream accuracy, and our proposed pred-intuit variant tends to make more accepted proposals than intuit-pred variants. Our choice of $k=5\%$ of $D$ appears reasonable.}
 \label{fig:ablation}
\end{figure}

We repeat the same experiment as shown in Table~\ref{tab:baselines}, but exploring different variants of our approach, and show the downstream accuracy plotted against the concept accuracy on the test set, as well as the number of accepted features for each variant in Figure~\ref{fig:ablation}.

\textbf{The combination of predictiveness and intuitiveness to generate the proposal is important for downstream accuracy.} Almost all variants that combine $\texttt{score}^{\texttt{intuit}}$ and $\texttt{score}^{\texttt{pred}}$ to make proposals outperform the pred and intuit variants on downstream accuracy. The left column of Figure~\ref{fig:ablation} shows that all of the intuit-pred variants, and 3 of the 4 pred-intuit variants outperform pred and intuit (the 4th performs similarly). The last variant has a high value of $k$, thresholding after the top 25\% of features. This is larger than it makes sense to set $k$ to in practice, but demonstrates that even with large values of $k$, results are no worse than using only `intuit' or `pred'.

\textbf{Both predictiveness and intuitiveness are also important for optimizing concept accuracy.} Our proposed variant and several others outperform pred and intuit in concept accuracy, and no variants perform substantially worse. In Yelp, our proposed variant, pred-intuit-5\% outperforms all other variants, and performs similarly or better to all variants in the Psych domain. This suggests that this variant is quite effective at optimizing both concept accuracy and downstream accuracy compared to other variants that use only $\texttt{score}^{\texttt{pred}}$ or $\texttt{score}^{\texttt{intuit}}$. We note that this particular setting of $k = 5\%$ of the total features was not chosen after hyperparameter optimization, but instead chosen a priori, however it seems to be a reasonable choice in both domains.

\textbf{Pred-intuit variants, including our proposed variant make more accepted proposals than intuit-pred variants, potentially engaging users more effectively.} The pred-intuit variants, including the one we propose, propose more accepted features than the intuit-pred variants. Both propose substantially more than the pred variant and less than the intuit variant. This is an important metric because even an approach that optimizes accuracy well will likely not be engaging to the user if it rarely makes relevant suggestions. Given that intuit has much lower downstream and concept accuracy, our proposed variant of pred-intuit-5\% balances these competing objectives well.

\section{Discussion}

Our approach offers advantages including providing the means to create a prediction system that allows people the ability to reason about the prediction process at a higher level of abstraction than the raw data, transparent concept definitions that can be validated for intuitiveness, and a feedback mechanism that allows for efficient use of human feedback. However it also has limitations that suggest interesting future work.

On our two tasks, all interpretable, concept-based approaches came at a cost to downstream accuracy. Future work could address this by making $g$ more flexible while still ensuring that its intuitiveness can be validated. Alternatively, this could be addressed by seeding our approach with concepts that are both intuitive and highly predictive in order to mitigate potential performance effects. The second option may also address cases where the user cannot easily specify a set of predictive concepts and an associated feature. 

While our results show good performance in realistic simulations of human feedback, we did not test it extensively with users. There remain important open questions including: how well can $g$ capture the high-level concepts in a variety of domains? And how accurately are people able to give the accept/reject feedback required by our algorithm? For the second, it may be possible that the user believes a feature-concept association is intuitive, but it actually makes the concept definition less representative of the user's concept. How realistic this scenario is is an interesting question for future work.

\section{Conclusion}

We proposed an approach for learning interpretable concept-based latent representations to extend interpretable machine learning methods to domains with uninterpretable features. We used human-in-the-loop training to learn transparent concept definitions that align with users' intuitive representation of a prediction problem, and we showed on real datasets with simulated concepts that our approach can learn representations that align substantially better with user-intuitive concepts. 

Transparent machine learning methods allow users to inspect system logic, potentially catching mistakes and improving models. Our approach scales these benefits to high-dimensional domains with unintuitive features without sacrificing transparency at the representation level. 

\section{Acknowledgments} This work was funded by NSF GRFP (grant no. DGE1745303) and the Sloan Fellowship. 

\small
\bibliographystyle{named}
\bibliography{bibliography}

\appendix

\section{Dataset}
\label{sec:dataset-appendix}

\subsection{Toy data}

We generated the toy dataset with 10,000 instances and split it 60\%/20\%/20\% train, valid, test. The pairs of correlated features were generated by first sampling a binary random variable $\sim$ Bernoulli(0.25), then for each of the 2 correlated features, flipping the variable with p=0.05. Labels were generated using the model described in Figure~\ref{fig:toy-example}

\subsection{Real datasets}

We used the Yelp dataset\footnote{https://www.yelp.com/dataset} from the Yelp dataset challenge. To process the data, we kept restaurants with at least 5 reviews, and used a bag of words feature representation, counting the number of times each word appears in all associated reviews for a restaurant. We then labeled as positive examples restaurants with star ratings $\geq 4$, and as negative examples restaurants with star ratings $\leq 2$, and subsampled the positive class to generate a class-balanced dataset. The words that we kept in the feature vectors occurred in reviews for between 10\% and 25\% of restaurants, allowing us to find features that were common enough to be useful predictors, but not so common that they were used for most restaurants. 

We used a dataset of patients from 2 New England hospitals with at least 1 MDD diagnosis (ICD9 codes 296.2x, 296.3x) or depressive disorder not otherwise specified (311), and without codes for schizophrenia, bipolar, and typical antipsychotics. Our prediction task was to determine whether the patient will be prescribed an atypical antipsychotic (Olanzapine, Quetiapine, Risperidone, Lurasidone, Aripiprazole, Brexpiprazole, Ziprasidone) within the year after their index antidepressant prescription. We subsampled negative examples to class balance the dataset. Feature vectors consist of counts of how often each ICD9, procedure and medication code are recorded for the patient in the 2 years preceding the index antidepressant prescription. We exclude codes that occur for less than 1\% of patients since there is a long tail of these codes that will not be highly predictive since they are recorded for few patients. We additionally remove numerical features from the dataset (patient age and date), and gender markers. We do this so we can use gender as a concept in our simulation studies that must be defined through proxies rather than through the recorded marker. In a real, clinical application, these features would be included in the dataset.

\textbf{Ground Truth Concept Definitions.} We defined a set of concepts and ground truth features associated with each concept. The concepts and features were generated by the first author, including input from domain experts in the Psych domain. Concepts were chosen to be reasonably predictive, and seed features were selected to be representative of the concept (based on the judgment of the first author, who compiled these lists of features associated with each concept). At each restart, concepts are seeded with one bolded term from each concept, such that $f$ generated with the seed features had large enough concept weights for each concept that adding new features to the concept definition affected the accuracy (this was sometimes not the case when a concept was seeded with a term that was irrelevant to the prediction task). Several features were excluded when they frequently lead to this condition being violated, suggesting that these were not predictive seed features. In practice, one would validate the predictiveness of the concepts and seed features before running this approach, which our choice of seed features mimics.

The concepts we define in the Yelp dataset are: `mention of service', `positive ambiance', and `positive food texture'. The associated words for each concept are listed below, with potential seed features in bold (concepts are seeded with a randomly chosen one of these):\\
`mention of service': management, manager, server, waiter, waitress, employee, hostess, cashier, bartender, orders, ordering, servers, register, refill, serves, serve, waitresses, refills, refill, \textbf{managers}, \textbf{reservation}, reservations, services, waiters\\
`positive ambiance': \textbf{cozy}, \textbf{ambience}, \textbf{ambiance}, welcoming, casual, friendly, music, modern, neighborhood, atmosphere, \textbf{comfortable}, quaint, \textbf{vibe}, comfort, comfortable, mood, welcome\\
`positive food texture': tender, \textbf{crispy}, crisp, \textbf{juicy}, \textbf{creamy}, moist, crunchy, \textbf{fluffy}, crunch\\

The concepts we define for the Psych dataset are: `anxiety' , `gender-female' , `hospital-ed' (hospitalization/emergency department visit) , `addiction'. The associated featuress for each concept are listed below, with potential seed features in bold:\\  
`anxiety': \textbf{30002}, \textbf{30000}, 30001, 7992, 3003, \textbf{rxnorm:2598}, \textbf{rxnorm:596}, rxnorm:6470, rxnorm:2353, rxnorm:3322, rxnorm:7781\\
`gender-female': v242, \textbf{c76801}, c59051, c58100, c76830, c76815, c76816, 6260, rxnorm:214559, v7610, 7210, 650, c76819, rxnorm:6691, c88142, c88141, v221, c59409, 6271, p7569, 6262, 64893, 6264, v103, 2189, p7534, c76805, v222, v7611, 6160, c59400, \textbf{c81025}, c82105, c76645, rxnorm:4100, 61610, v7231, v270, c76811, v163, rxnorm:214558, c88174, \textbf{drg:373}, 6202, rxnorm:384410, rxnorm:6373, \textbf{c59025}, 6253, c88175, 1749, 6221, 6259, 6268, 6272, 6289, 79380, 7950, 79500, c76090, c76091, c76092, c77057, \textbf{v7612}, \textbf{v762}, c82670, 65963, rxnorm:324044, c84146, v220, rxnorm:4083, c76817\\
`hospital-ed': \textbf{c99232}, c99231, c99222, c99233, c99238, c99223, c99282, c99285, \textbf{c99284}, c99283, c99281, c99239, c99253, c99219, c99218, \textbf{c99221}, \textbf{zINPATIENT}, c99254, c99252\\
`addiction': 30400, \textbf{c80100}, \textbf{3051}, \textbf{30500}, 29181, 30390, 30590, c82055, 30490, rxnorm:6813, \textbf{rxnorm:7407}, c80101, v1582\\
Codes starting with `c' are CPT codes, codes starting with `rxnorm' are medication codes, and the rest are ICD9 codes.

\section{Hyperparameters}
\label{sec:hyperparameters-appendix}

\subsection{Our Approach} We train $f$ using the scikit-learn implementation of logistic regression with no regularization term \citep{pedregosa2011scikit}.

\subsection{Interactive Concept-Learning Baselines}

For both interactive concept-learning baselines, we train the same $f$ (unregularized logistic regression implemented in scikit-learn \citep{pedregosa2011scikit}) as we do for our method. We use the concept probabilities directly in the downstream classification. This gives these approaches an advantage over our model and makes them slightly less interpretable, since our concepts are always constrained to be in $\{0,1\}$. 

\textbf{Concept Classifiers} Inspired by \citet{amershi2009concept}, we tune a set of concept-classifiers using concept labels, where the classifiers are l1-penalized logistic regressions so as to be simulatable. We request labels for the example that most improves the downstream accuracy of the model after retraining from a random subset of examples (while we use ground-truth concept labels in our simulation experiments, these would need to be estimated in practice). We search over a random subset of 100 examples to consider labeling. While searching over more examples will likely improve performance of the approach, it also increases the running time, which can seriously impact user experience in an interactive system. We generate the initial set of labels for each concept by labeling as positive examples of the concept all examples that have the seed term for the concept and randomly choosing 1 negative example of the concept to label. This gives the approach a roughly equivalent starting amount of information to our approach, which requires a seed term.

To determine the weight of the l1 regularization, we do grid search over a set of values to find the weight that produces final concept models with approximately the same number of features as our method has features associated with any concept. In practice, we search over values in the range 0.01 to 0.1 (after predetermining that the weight lies in this range), taking steps of size 0.01. We z-score the features before using this approach. 

\textbf{Topic Model} We also compare to the method in \citet{lund2018classifier} that tunes supervised topics through a set of curated anchor words to use in downstream prediction tasks. To make the interactions comparable to our approach, we propose a new anchor word as the highest probability word for the topic that is not already an anchor word, or a downstream label. We add rejected proposals to a set of ``irrelevant concepts'' not used in prediction since topic models must model all of the data--a feature not shared by our approach.

We run two variants of the topic-model approach in our experiments that create the ``irrelevant topics'' in two ways: in the first variant, we seed the model with 5 times as many non-concept-related topics as concept-related topics. We start with 1 topic for each concept, and each time we reject a term, we create a new topic with that word as the anchor word. Before adding rejected features to a new concept, we verify that they do not belong to the lists of related features for any other concepts. If they do, we ignore them since we do not want to prevent them from being suggested for the correct topic (although this would not be doable in practice). We call this TM, and report these results in the main body of the paper; they are better than TM-2. 

In the 2nd variant, TM-2, we generate anchor words for these by, for each new topic, taking the word that is the furthest from the existing anchor words using the Jaccard distance metric. We then assign rejected words to these topics by taking the topic with the closest anchor word to the rejected word based on Jaccard distance. These two variations allow us to explore whether pre-seeding the model with these ``irrelevant topics'' and allowing it to learn topics that more accurately correspond to our desired concepts from the beginning, or if creating ``irrelevant topics'' to specifically capture things that may be confused with our desired concepts is more effective. For both variants, we use $5*C$ irrelevant topics. 

We infer the topics by drawing a small number samples (specifically 10) of the topic vectors as suggested in \citet{lund-etal-2017-tandem} and computing probabilities by normalizing. We then binarize these to compute concept accuracy by taking all topics where the probability is greater than 0.1 for the document as 1 and the other topics as 0. Note that we train the logistic regression model for downstream prediction only on the topics that correspond to our desired concepts. 

\subsection{Non-Interactive Baselines}

The random forest model has 200 estimators, and we tune the maximum depth of the trees over the range $[5 , 10 , 25 , 50 , 100 , \text{None}]$ using 5-fold cross validation. The neural network has 1-hidden layer with the same number of hidden nodes as our approach has concepts--this is the comparable architecture to our approach. We use a sigmoid activation function and ADAM as an optimizer, and search over step sizes from $[ 0.001 , 0.01 , 0.1 , 1. ]$ using 5-fold cross-validation. We use batch size 32 and run it for 1000 iterations. For LR, we choose the l1-penalty that produces the closest number of non-zero coefficients to $C$ (the number of concepts). We search over values in the range 0.0001 to 1., taking steps of size 0.0001 between 0.0001 and 0.001, of size 0.001 between 0.001 and 0.01, etc. 

We z-score the features before using any of these approaches. We trained the random forest model, and the logistic regression models using the scikit-learn implementations \citep{pedregosa2011scikit}.

\section{Results}

\begin{figure}[h!]
 \centering
 \includegraphics[width=0.5\textwidth]{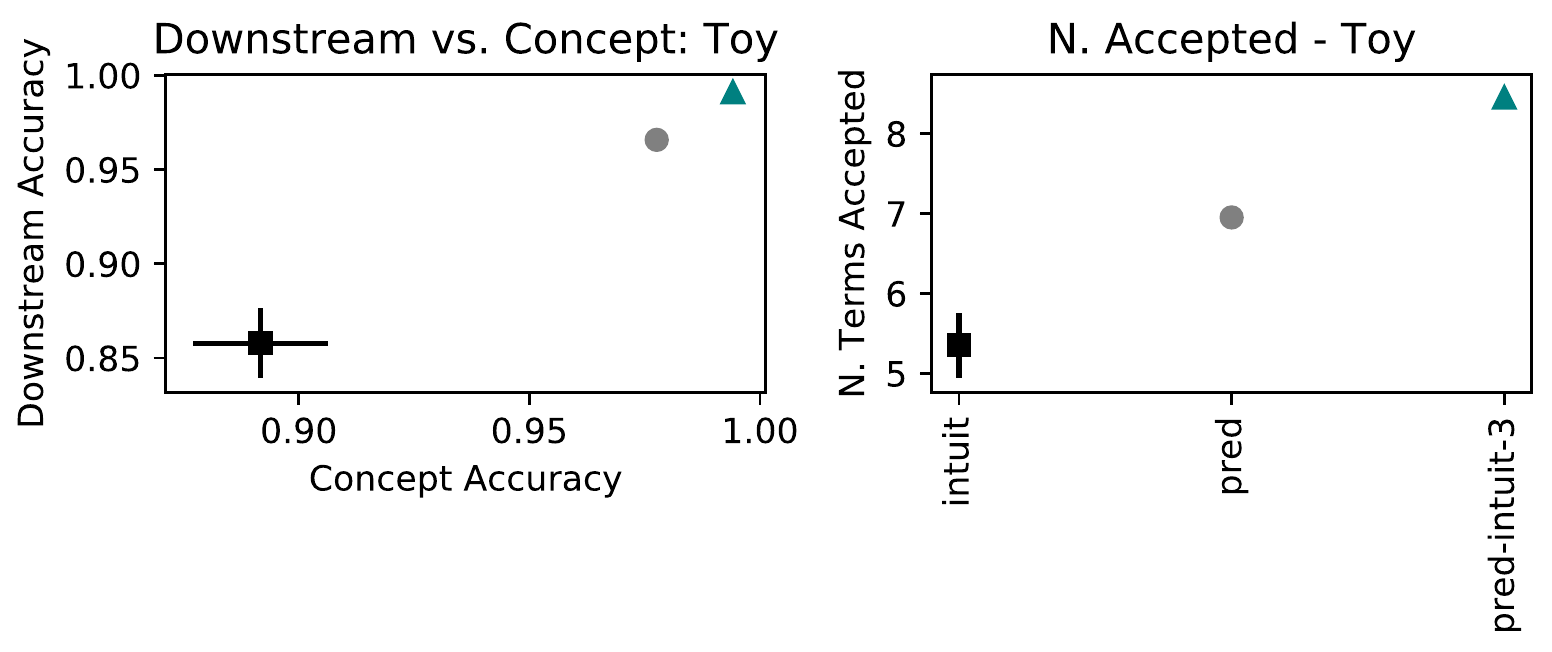}
 \caption{Downstream vs. concept accuracy on the test set for our proposed variant, the pred variant, and the intuit variant, as well as the number of accepted proposals made by each variant for 20 random restarts. Our variant works best by all metrics, followed by pred, then intuit.}
 \label{fig:toy-results}
\end{figure}

\begin{figure}[h!]
 \centering
 \includegraphics[width=.45\linewidth]{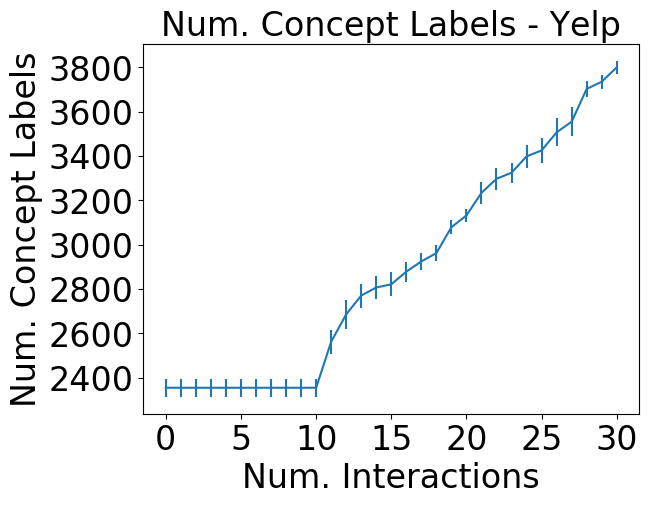}
 \includegraphics[width=.45\linewidth]{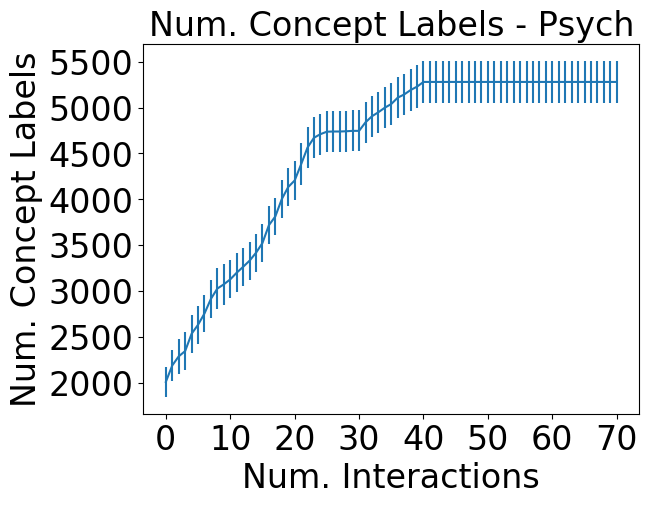}
 \caption{Number of positive concept labels in test set by number of proposals for our method. Yelp on left, psych on right. The number of positive concept labels substantially increases suggesting our method expands coverage.}
 \label{fig:coverage}
\end{figure}

\setlength{\tabcolsep}{3pt}
\begin{table}[h!]
\centering
\small
 \begin{tabular}{||l|r r|r r||} 
 	\hline
 	& \multicolumn{2}{|c|}{Yelp} & \multicolumn{2}{|c||}{Psych} \\
	Variant & Downstream & Concept & Downstream & Concept \\
	\hline\hline
	TM-2 & 0.71$\pm$0.01 & 0.61$\pm$0.01 & 0.61$\pm$0.00 & 0.64$\pm$0.00	\\ \hline
 \hline
 \end{tabular}
 \caption{Downstream accuracy and concept accuracy $\pm$ standard errors for the TM-2 variation of the anchor topic modeling baseline. These results are worse than the TM variant reported in the main text.}
 \label{tab:tm-variant}
\end{table}
\setlength{\tabcolsep}{6pt}

\textbf{Toy Example: Our approach achieves perfect downstream and concept accuracy while pred and intuit do not.} Figure~\ref{fig:toy-results} shows the downstream accuracy plotted against the concept accuracy as well as the number of accepted features for the pred, intuit and our proposed variant in the toy example. Our method works best, achieving perfect downstream and concept accuracy in 4 proposals per concept, followed by pred, then intuit. Each random restart corresponds to re-sampling the dataset and re-running the algorithms.

\textbf{Inclusion and Coverage: Our approach increases positive concept labels substantially, implying improved coverage.} Figure~\ref{fig:coverage} shows the number of positive concept labels by number of user interactions from the experiment above. In both domains, the number of positive concept labels grows substantially, and more than doubles from the seed features in the Psych domain. This has implications for fairness and robustness as it allows for multiple synonymous ways of coding for different concepts that capture different instances. In our model, these can all be recognized instead of relying on a single common coding as would likely be the case in a model constrained only to be sparse (without concepts, like LR).

\textbf{The TM-2 variant has worse performance than TM. }Results for the second variation of TM are in Table~\ref{tab:tm-variant}. Across the board, these are the same or worse than the TM results reported in Table~\ref{tab:baselines}, so we only compare TM to our approach.

\end{document}